\documentclass{egpubl}
\usepackage{eurovis2024}

\EuroVisShort  

\usepackage[T1]{fontenc}
\usepackage{dfadobe}

\usepackage{cite}  
\BibtexOrBiblatex
\electronicVersion
\PrintedOrElectronic
\ifpdf \usepackage[pdftex]{graphicx} \pdfcompresslevel=9
\else \usepackage[dvips]{graphicx} \fi

\usepackage{egweblnk}

\usepackage{booktabs}
\usepackage[normalem]{ulem}
\usepackage{graphicx}
\usepackage[rightcaption]{sidecap}

\usepackage{subcaption}
\useunder{\uline}{\ul}{}
\title[EG \LaTeX\ Author Guidelines]%
      {Evaluating Pre-Training Bias on Severe Acute Respiratory Syndrome Dataset}
      
\author[D. Rodrigues]
{\parbox{\textwidth}{\centering D.\,D. Rodrigues
\orcid{0000-0002-9544-117X}
}
        \\
{\parbox{\textwidth}{\centering $^1$Institute of Informatics, Federal University of the Rio Grande do Sul (UFRGS), Porto Alegre, RS, Brazil
       }
}
}

\begin{document}


\maketitle
\begin{abstract}
Machine learning (ML) is a growing field of computer science that has found many practical applications in several domains, including Health. However, as data grows in size and availability, and the number of models that aim to aid or replace human decisions, it raises the concern that these models can be susceptible to bias, which can lead to harm to specific individuals by basing its decisions on protected attributes such as gender, religion, sexual orientation, ethnicity, and others. Visualization techniques might generate insights and help summarize large datasets, enabling data scientists to understand the data better before training a model by evaluating pre-training metrics applied to the datasets before training, which might contribute to identifying potential harm before any effort is put into training and deploying the models. This work uses the severe acute respiratory syndrome dataset from OpenDataSUS to visualize three pre-training bias metrics and their distribution across different regions in Brazil. A random forest model is trained in each region and applied to the others. The aim is to compare the bias for the different regions, focusing on their protected attributes and comparing the model's performance with the metric values.
\end{abstract}  
\section{Introduction} 
Machine learning is an increasingly popular field in computer science that has attracted lots of interest from students and professionals. As the field's popularity increases, the number of available data to train the models also grows. Particularly in health, where data is central to the decision-making process, we have witnessed a growing increase in the application of ML algorithms to assist in diagnosis definition, prognosis, or treatment \cite{ghassemi2020review, miotto2018deep}. 

Machine learning in health also draws attention to the ability to inflict harm on certain groups or individuals due to erroneous pre-made assumptions. Suppose the act of training a machine learning model incorporates biased data; in that case, the model might exclude or favor certain classes, increasing the disparity for already under-represented and marginalized groups and systematically worsening already existing inequities in healthcare. \cite{Aquinojme-2022-108850}

 For instance, \cite{obermeyer2019dissecting} analyzed an algorithm used by US Hospitals to allocate patients to healthcare units and identified that the algorithm systematically discriminates against black people. The authors concluded that the algorithm was less likely to refer black people than white people who were equally sick to get appropriate help. Lower socioeconomic status was also associated with worse predictive model performance in the study conducted by \cite{juhn2022assessing}. According to the authors, this may be because people with lower socioeconomic status usually have less comprehensive medical registers due to less access to healthcare resources.

 This work aims to provide analysis and visualization of three pre-training metrics across the five regions in Brazil, visualizing the value for such metrics and training machine learning models with the data for each region. We aim to show that visualization techniques might aid researchers in bias identification on the data before any efforts to train the model, easing the model development and generating fair models.

\section{Theoretical Background}
In this chapter, we present an overview of the definition of bias in machine learning and go over the definition and formula for the three pre-training metrics used in this work.
\subsection{Bias in Machine Learning}
Bias is the inclination to think that some people or groups have advantages over other groups, leading to unfair treatment of some individuals. Data or algorithms can introduce bias in machine learning models. These biased models may make predictions based on protected attributes, which are sensitive information that could indicate prejudice against certain groups, such as race, sex, or age. In this work, we will focus on bias from data, focusing on two protected attributes on the datasets: sex and race.

\subsection{Pre-training Metrics}
This work uses three pre-training metrics to quantify the risk of bias in the datasets. The definition and formulation from the metrics were extracted from \cite{Hardt2021}. This section gives an overview of each metric and follows this notation:
\begin{itemize}
    \item facet $a$ represents the feature value that defines a demographic that bias favors (\emph{i.e.,} the overrepresented or advantaged group)
    \item facet $d$ represents the feature value that defines a demographic that bias disfavors (\emph{i.e.,} the underrepresented or disadvantaged group)
\end{itemize}

\subsubsection{Class Imbalance}
Class Imbalance measures the discrepancy of values in each class of the protected attribute. In the equation~\ref{eqn:CI_equation}, $n_a$ represents the number of values in facet $a$, \emph{i.e.,} the feature value that defines a demographic that bias favors (the advantaged group). Moreover, $n_d$ represents the number of values in facet $d$, \emph{i.e.,} the feature value that defines a demographic that bias disfavors (the disadvantaged group).

CI values range from -1 to 1. Positive values mean the facet $a$ has more representation than facet $d$. An ideal CI value is near zero, meaning the dataset represents both groups equally.

\begin{equation} 
\label{eqn:CI_equation}
 CI = (n_a - n_d)/(n_a + n_d)
\end{equation}

\subsubsection{Kullback-Leibler (KL) Divergence}
Kullback-Leibler (KL) Divergence measures the divergence between the label distributions for two facets, $a$ and $d$, represented by $P_a(y)$ and $P_d(y)$, respectively. The most commonly used implementation involves natural logarithms, meaning the metric measurement is in nats. The formula for this metric is in Equation~\ref{eqn:KL_Equation}.

\begin{equation} 
\label{eqn:KL_Equation}
 KL(P_a || P_d) = \sum_Y P_a(y)*log[P_a(y)/P_d(y)]
\end{equation}

\subsubsection{Kolmogorov-Smirnov (KS)}

Kolmogorov-Smirnov (KS) is equal to the maximum divergence between labels in the distribution for different facets of a dataset. It finds the most unbalanced label. This metric can also be used for multi-categorical targets since the number of terms in the equation is directly related to the number of possible values for the predicted attribute $y$. The formula for this metric is in Equation~\ref{eqn:KS_formula}, where $P_a(y)$ is the number of members in facet $a$ and $P_d(y)$ refers to the members in facet $d$. For this metric, the bigger the value, the more present the class disparity is, so ideally, it should result in a number near zero.

\begin{equation} 
\label{eqn:KS_formula}
 KS = max( |P_a(y) - P_d(y)| )
\end{equation}


\section{Related Work}
In recent years, developers have created several tools to address bias in machine learning models. For instance, \cite{Bellamy2018,Zehlike2017,Adebayo2016,tramer2015fairtest} are some tools that provide bias evaluation through different metrics based on the trained model's predictions and provide feedback as to what can be improved in the models to make fair predictions.

Numerous papers also focused on analyzing the bias impact on the protected attributes, such as \cite{Alelyani2021}, that used a general dataset with information about the census to detect whether bias might lead to unfair predictions for women and non-white people. \cite{Mandhala2022} has also used three datasets, evaluated the pre-training metrics, and then applied data augmentation techniques to balance the data while reducing disparity. Furthermore, \cite{SONG2019} analyzes the limitations of ML algorithms for predicting juvenile recidivism, comparing the performance and fairness from the results among ML models and SAVRY (Structured Assessment of Violence Risk in Youth). 

\cite{Comba2020} utilized visualization techniques to analyze COVID-19 data and demonstrated how such techniques can aid in comprehending different aspects of the pandemic. \cite{7466736} has also focused on how to apply visualization while discussing challenges and opportunities. While we can find many works on identifying and mitigating bias in the literature, there is a gap regarding data visualization tools and applying such techniques to help visualize and mitigate data disparities.
\section{Methodology}
We divide this work into the following steps: collect the data from OpenDataSus, pre-process the data to remove unwanted features (redundant and irrelevant features), split the data according to each region (North, Northeast, Central-West, Southeast, and South), evaluate the pre-training metrics for each region, train a Random Forest model on each region with an artificial task, infer for each region with the model from each region, respectively and report the performance for each model and compare it with the metrics.

\subsection{Data Collection}
This work uses three datasets from OpenDataSUS' Severe Acute Respiratory Syndrome (SRAG) from 2021, 2022, and 2023. Each row from the dataset contains medical data from one patient filled out at the health center by one physician. Some features were removed during the pre-processing as they were irrelevant to the context and might add noise to the training process. The datasets had, respectively, 762580, 379657, and 164356 rows (since in 2021 and 2022 the pandemic was far more active, it is expected to have more data from the health centers).
\subsection{Training and Performance Evaluation}
As we split the data for the five regions, we create two artificial tasks for the datasets: for 2021, we remove the ICU column and create the task to predict whether that patient will be admitted to ICU or not, and for 2022 and 2023, we remove the vaccination column and create the task to predict whether this patient is vaccinated against COVID-19 or not (one or more doses). We then train a Random Forest model with 300 estimators using the Gini index to split the nodes for each region. We then apply the five models to the five regions, resulting in 25 values for each performance metric. We report accuracy and F1-Score as performance metrics, and we focus our efforts on analyzing the false positives and false negatives for the different protected attribute classes. To have a fair visualization of the results, we normalized the values by dividing the true and false positives and negatives by the number of records in each class.

\section{Results}
In this section, we will go over the main results for the three analyzed years. The full reports along with the reproducible code were made available on GitHub \cite{Dimer_Evaluating_Pre-Training_Bias}
\subsection{2021}
\begin{figure}[htb]
    \centering
        \includegraphics[width=.8\linewidth]{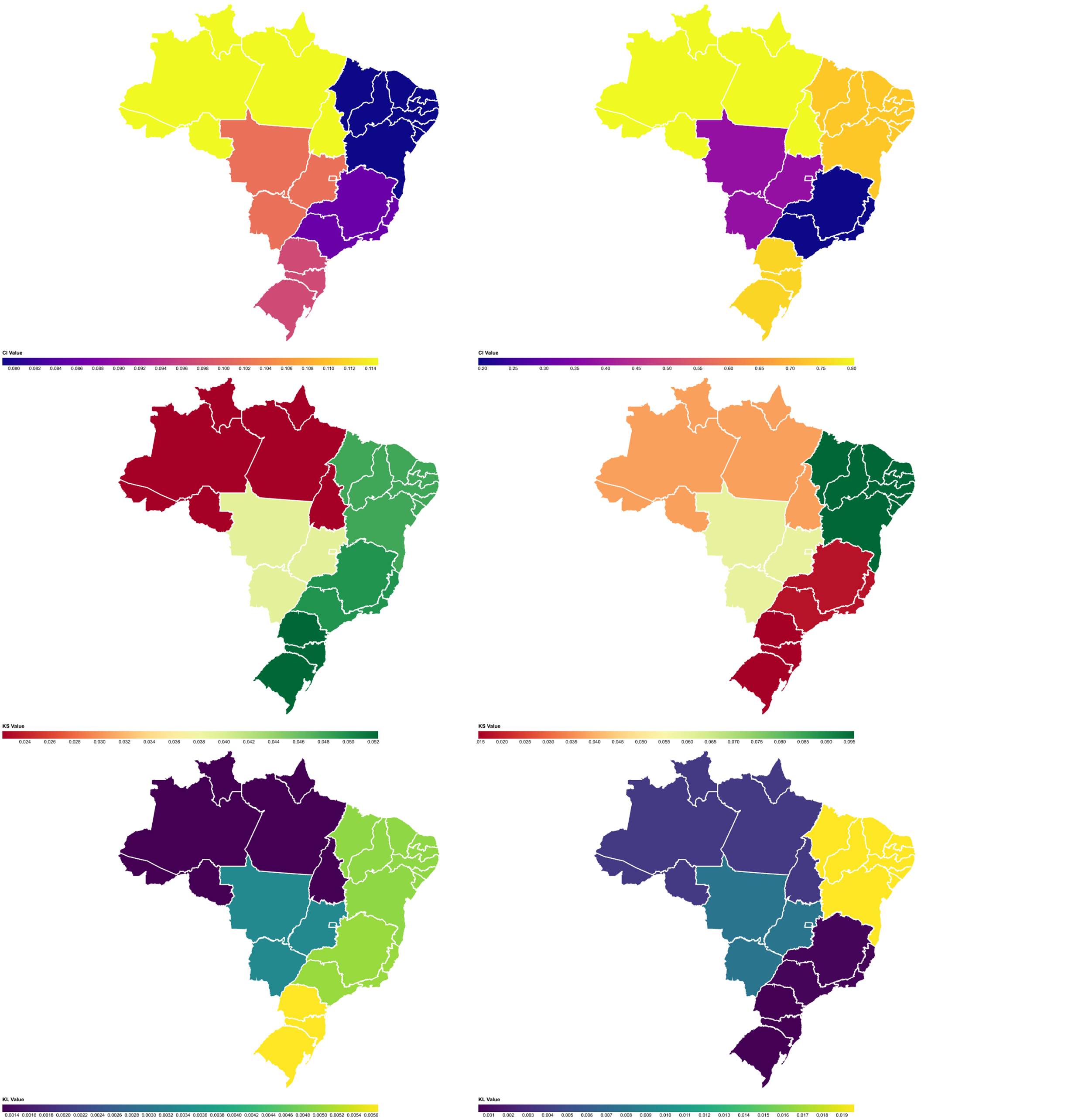}    
    \caption{\label{fig:chloro-2021}Choropleth maps for 2021, for Sex (right) and Race (left)}
\end{figure}

\begin{figure}[htb]
    \centering
        \includegraphics[width=.8\linewidth]{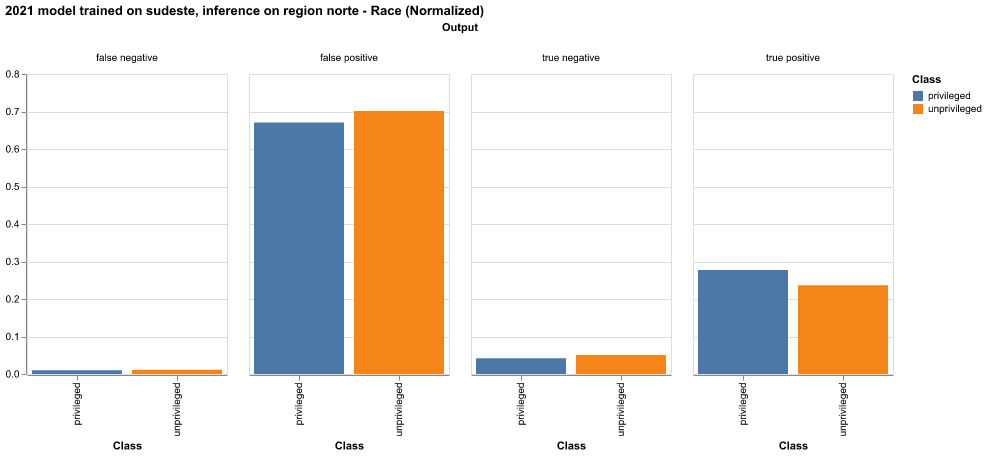}    
    \caption{\label{fig:example-2021-race}Prediction analysis for region north using a model trained on the southeast, for 2021}
\end{figure}

\begin{figure}[htb]
    \centering
        \includegraphics[width=.8\linewidth]{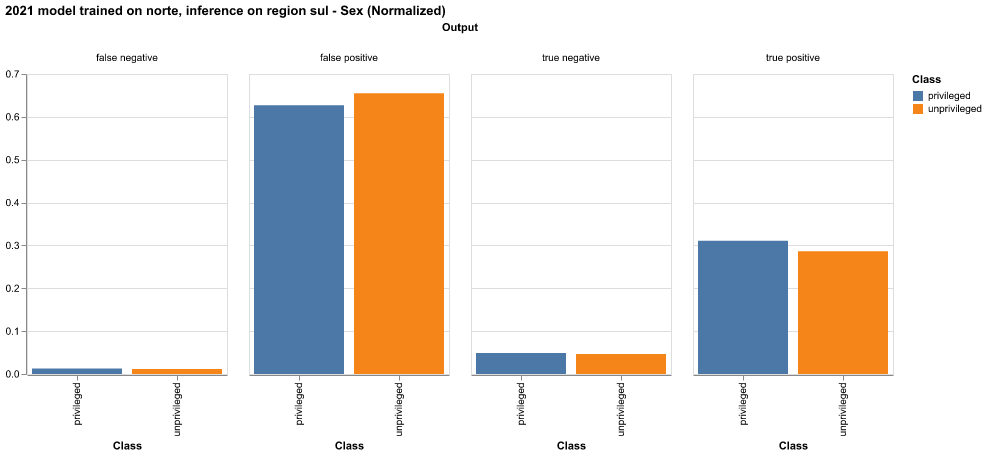}    
    \caption{\label{fig:example-2021-sex}Prediction analysis for region south using a model trained on the north, for 2021}
\end{figure}

For 2021, KL and KS metrics values raged from 0.0014 to 0.0056 and from 0.024 to 0.052 for \textbf{sex} and from 0.001 to 0.019 and from 0.015 to 0.095 for \textbf{race}, respectively. CI for sex ranged from 0.080 to 0.114, while race ranged from 0.20 to 0.80. The complete choropleth maps for the metrics are in Figure ~\ref{fig:chloro-2021}. For this year, the model's performance was considerably low, with predictions for the region northeast being the highest value (around 0.4 for all models, being the higher value from all regions for all models) and the worst performance predicting for the region north (around 0.30 for all models). These values could mean the task was too complex for the available data or the model was not general enough. Regarding the protected attributes, the model's performance was very similar for the privileged and unprivileged classes, not having a visible disparity in the predictions for these classes and having a more pronounced false positive and true positive rate. An example of the comparison between the two classes is in Figure~\ref{fig:example-2021-race} for \textbf{race} using the model from southeast on data from region north and Figure~\ref{fig:example-2021-sex} for \textbf{sex} using the model from north on region south.

\subsection{2022}
\begin{figure}[htb]
    \centering
        \includegraphics[width=.8\linewidth]{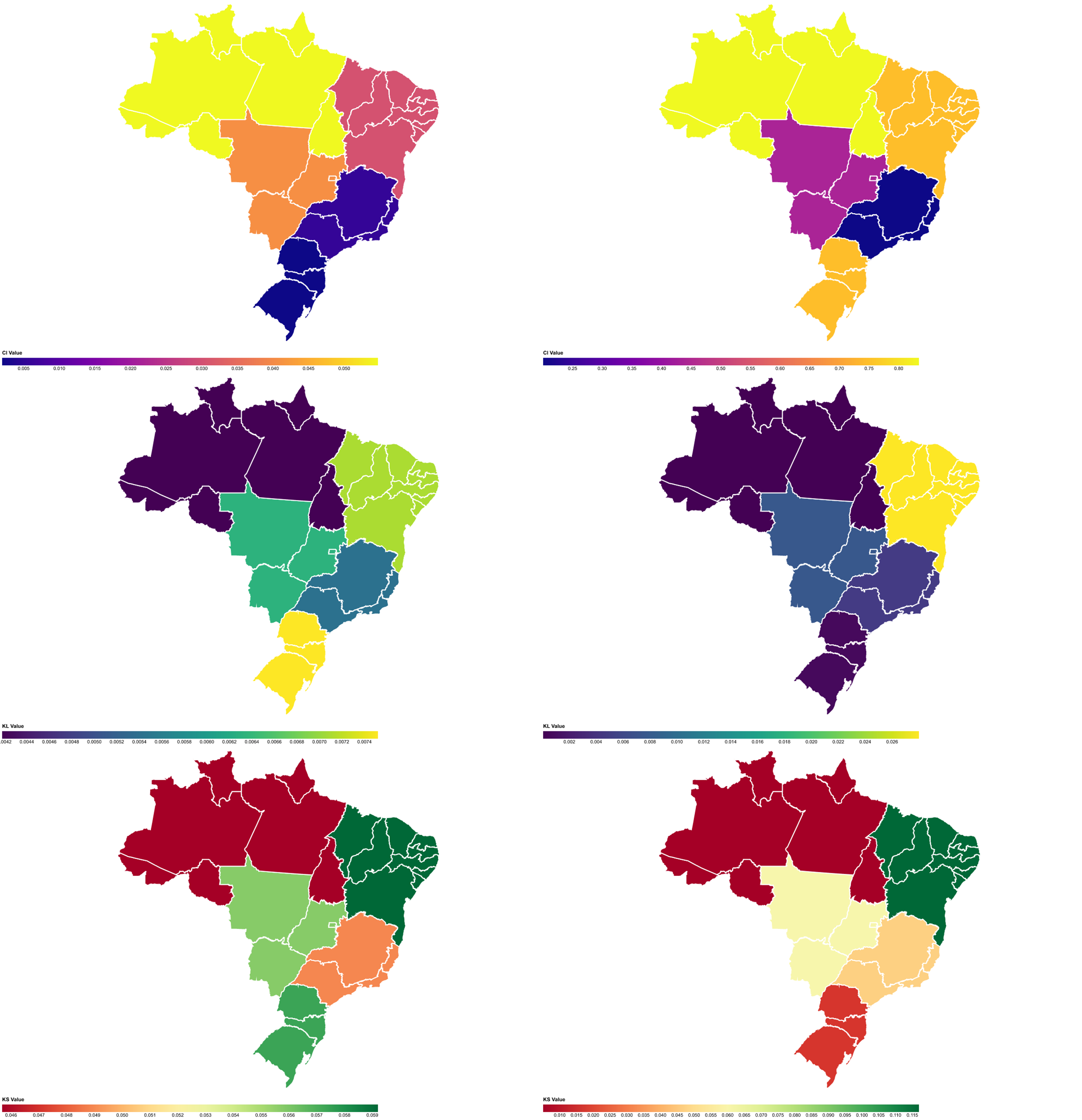}    
    \caption{\label{fig:chloro-2022}Choropleth maps for 2022, for Sex (right) and Race (left)}
\end{figure}

\begin{figure}[htb]
    \centering
        \includegraphics[width=.8\linewidth]{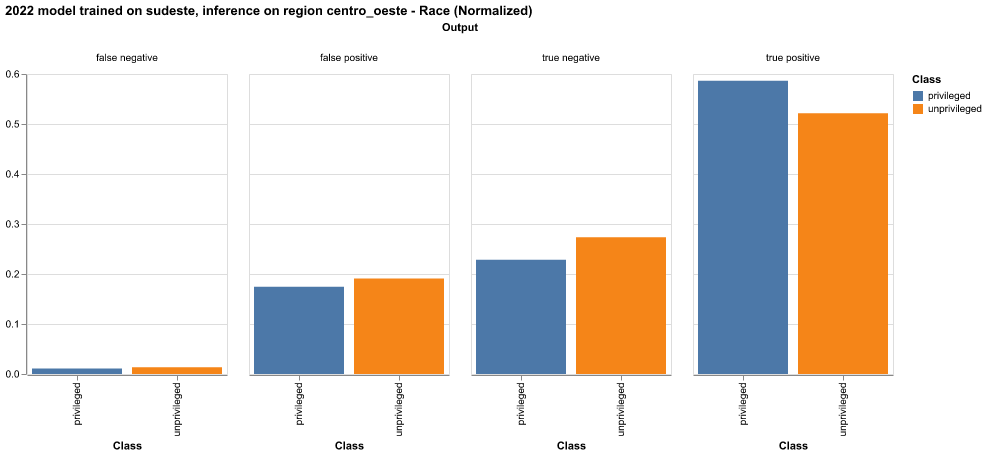}    
    \caption{\label{fig:example-2022-race}Prediction analysis for region midwest using a model trained on the southeast, for 2022}
\end{figure}

\begin{figure}[htb]
    \centering
        \includegraphics[width=.8\linewidth]{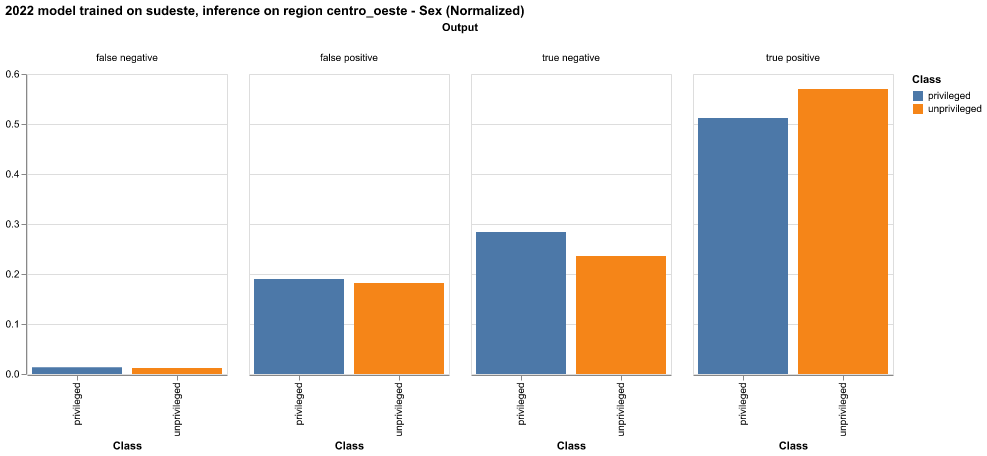}    
    \caption{\label{fig:example-2022-sex}Prediction analysis for region midwest using a model trained on the southeast, for 2022}
\end{figure}

In the second dataset, for 2022, KL and KS metrics values ranged from 0.0042 to 0.0074 for \textbf{sex}, from 0.046 to 0.059 for \textbf{race}, from 0.002 to 0.026 for \textbf{sex} and from 0.010 to 0.115 to \textbf{race}, respectively. For CI on \textbf{sex}, the value ranged from 0.005 to 0.050; in \textbf{race}, it ranged from 0.25 to 0.80. The full choropleth maps are in Figure~\ref{fig:chloro-2022}. The performance this year was considerably high, with the south region having the higher accuracy and F1-score across all five models (around 0.86 accuracy and 0.90 F1-score) and the other regions varying between 0.85 (southeast) and 0.75 (north). Regarding the protected attributes, we do not see a significant difference in the predictions for the unprivileged and privileged classes, with most errors coming from false positives. An example of the predictions is in Figure~\ref{fig:example-2022-sex} for \textbf{sex} and in Figure~\ref{fig:example-2022-race} for \textbf{race}. 

\subsection{2023}
\begin{figure}[htb]
    \centering
        \includegraphics[width=.8\linewidth]{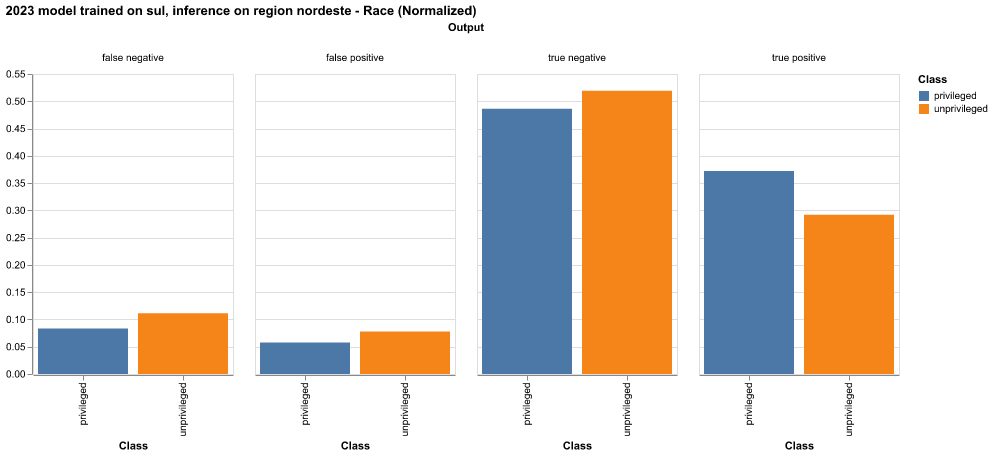}    
    \caption{\label{fig:example-2023-race}Prediction analysis for region northeast using a model trained on the south, for 2023}
\end{figure}

\begin{figure}[htb]
    \centering
        \includegraphics[width=.8\linewidth]{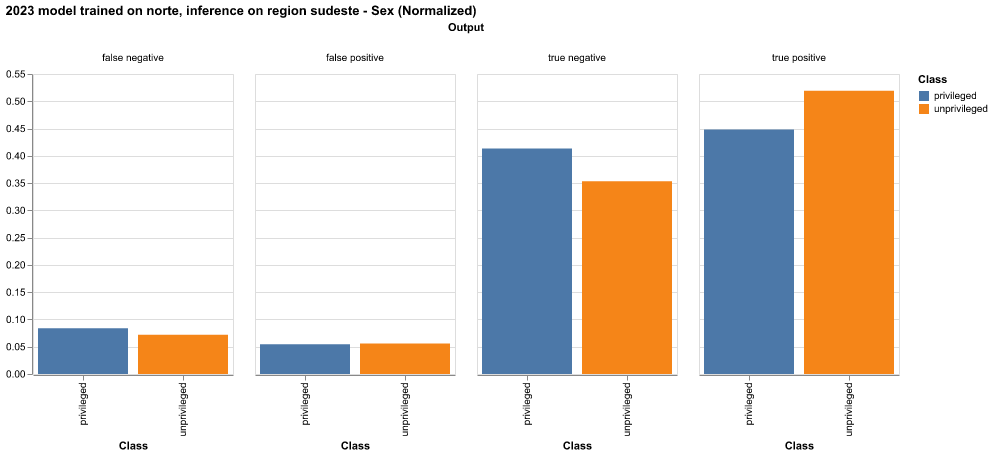}    
    \caption{\label{fig:example-2023-sex}Prediction analysis for region southeast using a model trained on the north, for 2023}
\end{figure}

In the third dataset, for 2023, KL and KS ranged from 0.0010 to 0.0105 for \textbf{race}, 0.0040 to 0.0125 for \textbf{sex}, from 0.020 to 0.070 for \textbf{race}, and from 0.044 to 0.080 to \textbf{sex}, respectively. CI ranged from 0.20 to 0.80 in \textbf{race} and from 0.028 to 0.076 in \textbf{sex}. The accuracy and F1-Score for this dataset were higher in every region the model corresponded to, with values above 0.96, with the other regions ranging between 0.81 and 0.85. Considering the protected attributes, there is a fair distribution of errors between false positives and false negatives, with no substantial discrepancy between the privileged and unprivileged classes. Examples from the prediction analysis are in Figure~\ref{fig:example-2023-race} for \textbf{race} and Figure~\ref{fig:example-2023-sex} for \textbf{sex}.

\begin{figure}[htb]
    \centering
        \includegraphics[width=.8\linewidth]{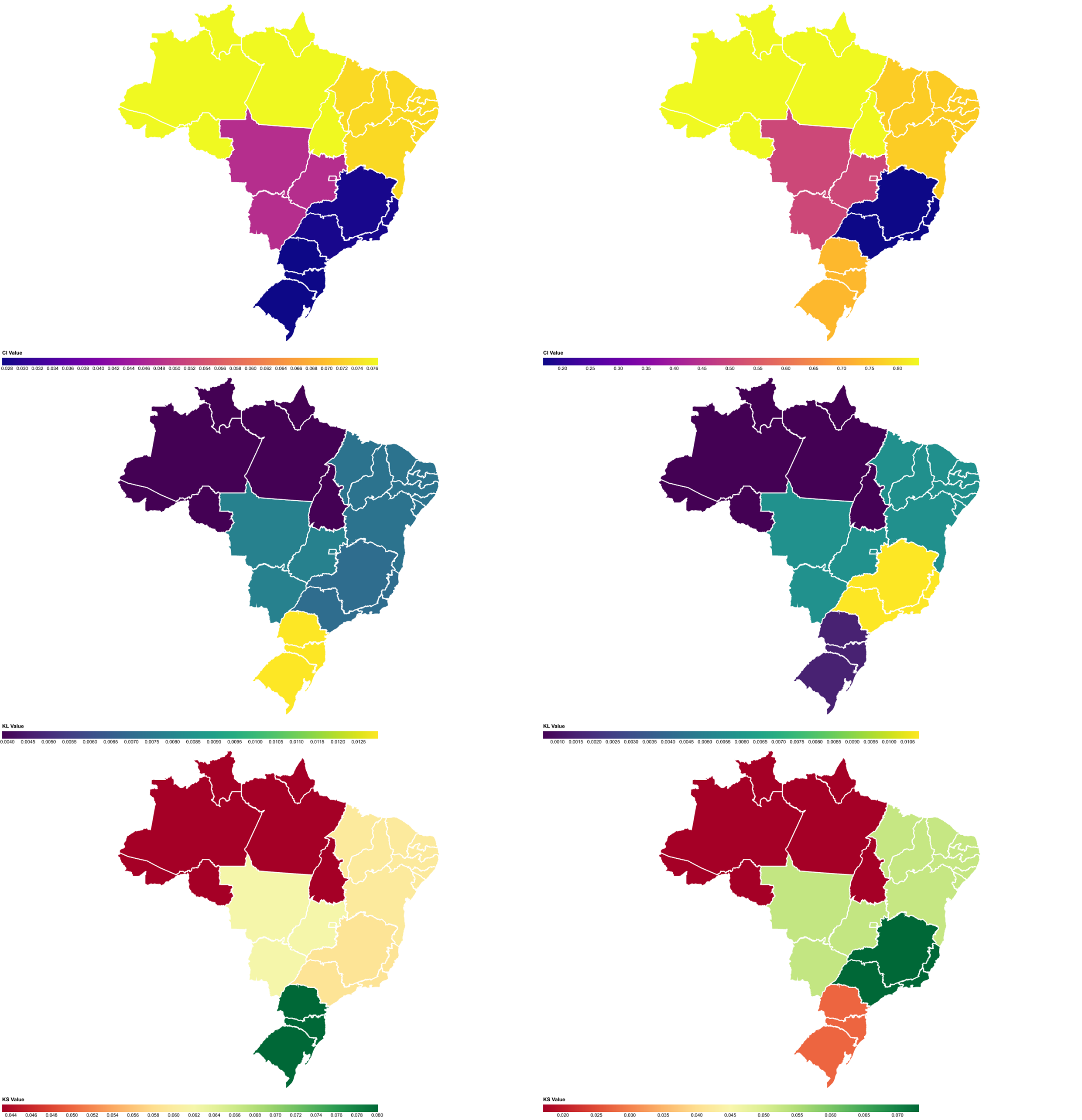}    
    \caption{\label{fig:chloro-2023}Choropleth maps for 2023, for Sex (right) and Race (left)}
\end{figure}

\section{Conclusion}
The visual analysis from the metrics values for these particular datasets did not indicate a high risk of bias in most metrics (KS and KL), as the values were too low. It might indicate that this particular data is not susceptible to bias, and therefore, the trained model would not discriminate against the analyzed groups (race and sex). For the Class Imbalance, it does not agree with the other two metrics, indicating a risk of bias that is not confirmed by the other metrics, pointing how important it is to use more than one metric to quantify the risk of bias. It does not mean, however, that if all the metrics values were high, we would see a discrepancy in the predicted values, as this was not observed in any case in this work.

Analyzing the race distribution in Brazil, we can infer that a model trained in a region where the majority is non-white would have poor performance in another region where the majority is white. However, as seen from the pre-training metrics, this does not reflect the actual scenario, where the performance was constant across the protected attributes. It could be because the pandemic hit equally different demographics, and the public health system in Brazil was effective in filling the notification sheet for all groups, which is a fair assumption since Brazil has a public health system that should be accessible to the entire population.

\bibliographystyle{eg-alpha-doi} 
\bibliography{egbibsample}

\newcommand{\etalchar}[1]{$^{#1}$}
\begin{thebibliography}{\uppercase{MWW{\etalchar{*}}18}}

\bibitem[ACH{\etalchar{*}}23]{Aquinojme-2022-108850}
\textsc{Aquino Y. S.~J., Carter S.~M., Houssami N., Braunack-Mayer A., Win K.~T., Degeling C., Wang L., Rogers W.~A.}:
\newblock Practical, epistemic and normative implications of algorithmic bias in healthcare artificial intelligence: a qualitative study of multidisciplinary expert perspectives.
\newblock \emph{Journal of Medical Ethics} (2023).
\newblock URL: \url{https://jme.bmj.com/content/early/2023/02/22/jme-2022-108850}, \href {http://arxiv.org/abs/https://jme.bmj.com/content/early/2023/02/22/jme-2022-108850.full.pdf} {\path{arXiv:https://jme.bmj.com/content/early/2023/02/22/jme-2022-108850.full.pdf}}, \href {https://doi.org/10.1136/jme-2022-108850} {\path{doi:10.1136/jme-2022-108850}}.

\bibitem[Ade16]{Adebayo2016}
\textsc{Adebayo J.}:
\newblock Fairml: Toolbox for diagnosing bias in predictive modeling.

\bibitem[Ale21]{Alelyani2021}
\textsc{Alelyani S.}:
\newblock Detection and evaluation of machine learning bias.
\newblock \emph{Applied Sciences (Switzerland) 11} (7 2021).
\newblock \href {https://doi.org/10.3390/app11146271} {\path{doi:10.3390/app11146271}}.

\bibitem[BDH{\etalchar{*}}18]{Bellamy2018}
\textsc{Bellamy R. K.~E., Dey K., Hind M., Hoffman S.~C., Houde S., Kannan K., Lohia P., Martino J., Mehta S., Mojsilovic A., Nagar S., Ramamurthy K.~N., Richards J., Saha D., Sattigeri P., Singh M., Varshney K.~R., Zhang Y.}:
\newblock Ai fairness 360: An extensible toolkit for detecting, understanding, and mitigating unwanted algorithmic bias.
\newblock URL: \url{http://arxiv.org/abs/1810.01943}.

\bibitem[Com20]{Comba2020}
\textsc{Comba J. L.~D.}:
\newblock Data visualization for the understanding of covid-19.
\newblock \emph{Computing in Science \& Engineering 22} (11 2020), 81--86.
\newblock \href {https://doi.org/10.1109/MCSE.2020.3019834} {\path{doi:10.1109/MCSE.2020.3019834}}.

\bibitem[Dim]{Dimer_Evaluating_Pre-Training_Bias}
\textsc{Dimer D.}:
\newblock {Evaluating Pre-Training Bias on Severe Acute Respiratory Syndrome Dataset}.
\newblock URL: \url{https://github.com/diegodimer/vis}.

\bibitem[GB16]{7466736}
\textsc{Gotz D., Borland D.}:
\newblock Data-driven healthcare: Challenges and opportunities for interactive visualization.
\newblock \emph{IEEE Computer Graphics and Applications 36}, 3 (2016), 90--96.
\newblock \href {https://doi.org/10.1109/MCG.2016.59} {\path{doi:10.1109/MCG.2016.59}}.

\bibitem[GNS{\etalchar{*}}20]{ghassemi2020review}
\textsc{Ghassemi M., Naumann T., Schulam P., Beam A.~L., Chen I.~Y., Ranganath R.}:
\newblock A review of challenges and opportunities in machine learning for health.
\newblock \emph{AMIA Summits on Translational Science Proceedings 2020} (2020), 191.

\bibitem[HCC{\etalchar{*}}21]{Hardt2021}
\textsc{Hardt M., Chen X., Cheng X., Donini M., Gelman J., Gollaprolu S., He J., Larroy P., Liu X., McCarthy N., Rathi A., Rees S., Siva A., Tsai E., Vasist K., Yilmaz P., Zafar M.~B., Das S., Haas K., Hill T., Kenthapadi K.}:
\newblock Amazon sagemaker clarify: Machine learning bias detection and explainability in the cloud.
\newblock URL: \url{http://arxiv.org/abs/2109.03285 http://dx.doi.org/10.1145/3447548.3467177}, \href {https://doi.org/10.1145/3447548.3467177} {\path{doi:10.1145/3447548.3467177}}.

\bibitem[JRW{\etalchar{*}}22]{juhn2022assessing}
\textsc{Juhn Y.~J., Ryu E., Wi C.-I., King K.~S., Malik M., Romero-Brufau S., Weng C., Sohn S., Sharp R.~R., Halamka J.~D.}:
\newblock Assessing socioeconomic bias in machine learning algorithms in health care: a case study of the houses index.
\newblock \emph{Journal of the American Medical Informatics Association 29}, 7 (2022), 1142--1151.

\bibitem[MBMK22]{Mandhala2022}
\textsc{Mandhala V.~N., Bhattacharyya D., Midhunchakkaravarthy D., Kim H.~J.}:
\newblock Detecting and mitigating bias in data using machine learning with pre-training metrics.
\newblock \emph{Ingenierie des Systemes d'Information 27} (2 2022), 119--125.
\newblock \href {https://doi.org/10.18280/isi.270114} {\path{doi:10.18280/isi.270114}}.

\bibitem[MWW{\etalchar{*}}18]{miotto2018deep}
\textsc{Miotto R., Wang F., Wang S., Jiang X., Dudley J.~T.}:
\newblock Deep learning for healthcare: review, opportunities and challenges.
\newblock \emph{Briefings in Bioinformatics 19}, 6 (2018), 1236--1246.

\bibitem[OPVM19]{obermeyer2019dissecting}
\textsc{Obermeyer Z., Powers B., Vogeli C., Mullainathan S.}:
\newblock Dissecting racial bias in an algorithm used to manage the health of populations.
\newblock \emph{Science 366}, 6464 (2019), 447--453.

\bibitem[TAG{\etalchar{*}}15]{tramer2015fairtest}
\textsc{Tramer F., Atlidakis V., Geambasu R., Hsu D., Hubaux J.-P., Humbert M., Juels A., Lin H.}:
\newblock Fairtest: Discovering unwarranted associations in data-driven applications.
\newblock \emph{arXiv preprint arXiv:1510.02377} (2015).

\bibitem[TMGC19]{SONG2019}
\textsc{Tolan S., Miron M., G\'{o}mez E., Castillo C.}:
\newblock Why machine learning may lead to unfairness: Evidence from risk assessment for juvenile justice in catalonia.
\newblock In \emph{Proceedings of the Seventeenth International Conference on Artificial Intelligence and Law} (New York, NY, USA, 2019), ICAIL '19, Association for Computing Machinery, p.~83–92.
\newblock URL: \url{https://doi.org/10.1145/3322640.3326705}, \href {https://doi.org/10.1145/3322640.3326705} {\path{doi:10.1145/3322640.3326705}}.

\bibitem[ZCB{\etalchar{*}}17]{Zehlike2017}
\textsc{Zehlike M., Castillo C., Bonchi F., Hajian S., Megahed M.}:
\newblock Fairness measures: Datasets and software for detecting algorithmic discrimination.
\newblock URL: \url{http://fairness-measures.org}.

\end{thebibliography}

\end{document}